\documentclass{sig-alternate-05-2015}[10pt]

\usepackage{epstopdf}
\usepackage{cite}
\usepackage{amsmath}
\usepackage{multirow}
\usepackage{epsfig}
\usepackage{mathrsfs}
\usepackage{amssymb}
\usepackage{comment}
\usepackage{array}
\usepackage{blkarray}
\usepackage{enumerate}
\usepackage{url}
\usepackage{chemarrow}
\usepackage{color}
\usepackage[tight,footnotesize]{subfigure}
\usepackage{fancyhdr}

\makeatletter
\newif\if@restonecol
\makeatother

\usepackage[lined,ruled,linesnumbered]{algorithm2e}
\makeatletter
\def\old@comma{,}
\catcode`\,=13
\def,{%
  \ifmmode%
    \old@comma\discretionary{}{}{}%
  \else%
    \old@comma%
  \fi%
}
\makeatother

\newtheorem{theorem}{Theorem}

\newtheorem{corollary}[theorem]{Corollary}

\newcommand{\prob}{\mbox{Pr}}

\DeclareMathAlphabet{\mathcal}{OMS}{cmsy}{m}{n}

\usepackage[tight,footnotesize]{subfigure}

\newfont{\mycrnotice}{ptmr8t at 7pt}
\newfont{\myconfname}{ptmri8t at 7pt}
\let\crnotice\mycrnotice%
\let\confname\myconfname%

\makeatletter
\def\@copyrightspace{\relax}
\makeatother

\begin{document}

\thispagestyle{fancy}
\lhead{\textit{Technical Report, IBM T. J. Watson Research Center, Yorktown, NY, USA, July, 2016.}}
\rhead{} 

\title{Query Answering via Decentralized Search}
\numberofauthors{1}
\author{
%
%
\alignauthor Liang Ma\\
       \affaddr{IBM T. J. Watson Research}\\
       \affaddr{Yorktown, NY, USA}\\
       \email{maliang@us.ibm.com}
}

\maketitle

\subsection*{Abstract}
Expert networks are formed by a group of expert-profes\-sionals with different specialties to collaboratively resolve specific queries posted to the network. In such networks, when a query reaches an expert who does not have sufficient expertise, this query needs to be routed to other experts for further processing until it is completely solved; therefore, query answering efficiency is sensitive to the underlying query routing mechanism being used. Among all possible query routing mechanisms, decentralized search, operating purely on each expert's local information without any knowledge of network global structure, represents the most basic and scalable routing mechanism, which is applicable to any network scenarios even in dynamic networks. However, there is still a lack of fundamental understanding of the efficiency of decentralized search in expert networks. In this regard, we investigate decentralized search by quantifying its performance under a variety of network settings. Our key findings reveal the existence of network conditions, under which decentralized search can achieve significantly short query routing paths (i.e., between $O(\log n)$ and $O(\log^2 n)$ hops, $n$: total number of experts in the network). Based on such theoretical foundation, we further study how the unique properties of decentralized search in expert networks is related to the anecdotal small-world phenomenon. In addition, we demonstrate that decentralized search is robust against estimation errors introduced by misinterpreting the required expertise levels. To the best of our knowledge, this is the first work studying fundamental behaviors of decentralized search in expert networks. The developed performance bounds, confirmed by real datasets, are able to assist in predicting network performance and designing complex expert networks.\looseness=-1

\section{Introduction}
\label{intro}

Expert networks are composed of a group of expert-pro\-fessionals, who cooperate with each other to solve specific queries (e.g., reported by clients) using their professional knowledge in a variety of related subjects. Such expert networks are abundant in real life especially in commercial organizations, where networks with specialized experts are maintained to provide consulting/troubleshooting services. The collaboration among experts is knowledge-driven, manifesting in the process of expert searching: Upon receiving a query by an expert, she first attempts to solve the problem specified in the query; if she fails, then this query is routed to another expert for further processing. This process continues until the query is resolved. One canonical example of expert networks is the enterprise call center. Specifically, in a call center, if a query ticket cannot be solved by the first responding agent, then a series of processing/forwarding attempts are triggered until qualified agents are found. The fundamental goal regarding expert networks is to route each query to experts with sufficient expertise in a timely and accurate manner.
This is a challenging issue as efficient query routing mechanisms depend on the professional knowledge of each individual expert as well as the social knowledge of other experts' specialties possessed at each expert.
When the expert profiles (e.g., expertise) are not properly maintained in expert networks, finding the most knowledgeable experts with high probability while minimizing the number of routing steps is explored in \cite{Balog06SIGIR,Serdyukov08CIKM}. On the other hand, even if each expert's profile is accurately exposed to all of her contacts, this routing issue remains challenging. In particular, under the assumption that each expert has connections to only a limited number of other experts, a series of routing rules are proposed in \cite{Shao08KDD,Zhang12AAMAS} for improving the resolution efficiency in specific tasks (e.g., IT services). Under the same assumption, generative models \cite{Miao10KDD} are developed for making global expert recommendations by estimating all possible routes to potential resolvers. The efficacy of these mechanisms, however, highly relies on the broad knowledge of network global structure; in addition, these mechanisms, requiring non-negligible training periods, are generally complicated and sensitive to operational scenarios, thus not applicable to large-scale or dynamic networks (e.g., with experts joining/leaving the network). All these limitations, therefore, motivate us to consider if there exist simple yet efficient query routing solutions that function with only basic network information and are robust against network variations.\looseness=-1

Among all possible query routing mechanisms, \emph{decentralized search}, operating purely on each expert's local information, represents the most basic and adaptive routing mechanism in expert networks. Specifically, under decentralized search, before reaching experts with sufficient expertise for a given query, each intermediate expert forwards this query to one of her contacts with the highest problem solving abilities, which therefore forms a pure local-information-based forwarding rule. Since no historical training data or network global structure is required for routing decision making, decentralized search can be broadly adopted by any network scenarios for query routing. However, decentralized search, greedy in nature at each routing step, is generally ignored in the research community mainly for the following reason: When an expert forwards a query to one of her contacts via decentralized search, she has no clue whether this decision would successfully lead to a short path through the entire network, and thus the efficiency of decentralized search is uncertain. We note, however, without fundamental understanding such simple decentralized search, we can never justify the value/necessity of designing other complicated query routing mechanisms for expert networks. Therefore, in this paper, we consider this unsolved fundamental problem: What is the efficiency of decentralized search in expert networks? We study this problem by quantifying the performance of decentralized search under various network settings so as to understand under what conditions decentralized search can achieve efficiency/inefficiency in query routing.\looseness=-1

In this paper, the basic approach we employ to study the performance of decentralized search is to establish its performance bounds in generic expert network models. Such models should capture two main connection properties among experts: (i) experts are rich in connections to peer experts with similar expertise; (ii) each expert also tends to connect to a few experts with fairly dissimilar expertise. Integrating these two properties that characterize expert social connections, we propose two expert network models. In these models, local expert connections (experts with similar expertise) enable the formation of the basic network structure, on top of which long-range expert connections (experts with fairly dissimilar expertise) determine to what extent the expert inter-connections do not respect such basic network structure. We prove that the natural superposition of these two properties in expert networks can lead to high efficiency of decentralized search without any centralized guidance under a range of network settings. Accordingly, if an expert network is verified to satisfy such conditions that guarantee efficient decentralized search, then there is no need to design complicated query routing mechanisms as the light-weighted decentralized search will suffice. Furthermore, by a case study of real datasets, we demonstrate how commercial expert networks may take proactive actions to train their constituent experts, which equivalently approaches the efficient query routing conditions discussed in this paper.

To gain more insights into decentralized search, we then consider how its efficiency relates to the intriguing and pervasive \emph{small-world phenomenon} \cite{Milgram67smallworld,Milgram69,Korte70psp}, a principle stating that any two individuals in the network are connected by a short chain of intermediate acquaintances. Note that as opposed to observing small-world phenomenon in expert networks, we aim to study, using the above theoretical results, the relationship between the existence of short routing chains and the efficacy of decentralized search in finding such short chains, thus providing better understanding of the uniqueness of small-world navigation via decentralized search in expert networks.\looseness=-1

\subsection{Further Discussions on Related Work}

Regarding expert networks, most existing works \cite{Balog06SIGIR,Serdyukov08CIKM,Shao08KDD,Zhang12AAMAS} seek to develop/improve query routing mechanisms, where different levels of network global knowledge are required. With network historical data, \cite{Banerjee08SNAKDD} develops a Markov Decision Process (MDP) model to optimize routing policies. However, the correlation between successful query answering probability and each expert's expertise level is ignored in the proposed model. Employing game theory, \cite{Kleinberg05FOCS} proposes query incentive networks to understand agent collaborations and interactions in on-line communities. In addition, routing efficiency improvement is investigated in \cite{Zeng2006optimal} when additional expert contacts are carefully chosen. Our work belongs to a different but closely related line of work that focuses on the fundamental understanding of the most basic query routing mechanism in expert networks. Our work shares similar goals with \cite{Sun14KDD} in that \cite{Sun14KDD} tries to build models to understand routing behaviors in expert networks, particularly in human factors that influence routing tendencies. However, \cite{Sun14KDD} does not show how the efficiency of such routing behaviors are affected by network properties. By contrast, we not only present deep insights into decentralized search in expert networks, but also show how its efficacy is related to network's structural and social properties.

For the underlying influence of network properties on the routing efficiency, \cite{Bollobas82, Bollobas88SIAMA} show that every pair of nodes are joined by a path of length $O(\log n)$ ($n$: total number of nodes in the network) in a randomly generated graph. The existence of such short paths is maintained even when the network demonstrates certain structural properties\cite{Watts98Nature}. Further, more complicated statistical models are considered in \cite{Draief2006efficient} for node inter-connections (e.g., Poisson distributions). When the number of links incident to nodes follows the power-law distribution, \cite{Adamic01Phy} explores how such distribution may affect routing preference. In addition, with special network properties, networks may exhibit small-world phenomenon \cite{Milgram67smallworld,Milgram69,Korte70psp}. The intriguing characteristic of small-world phenomenon has stimulated numerous compelling research results \cite{Guare90SixDegrees,Watts98Nature,Kleinberg00STOC,Kleinberg00Nature,Kleinberg06ICM,Dietzfelbinger14SODA}, among which \cite{Kleinberg00STOC} is the first work showing that there exists one and only one network setting that enables efficient searching algorithms. In this paper, we do not seek to observe small-world phenomenon in another type of networks (i.e., expert networks); by contrast, we aim to understand the small-world phenomenon with characteristics that uniquely exist in expert networks using our fundamental theoretical results on decentralized search. To this end, we prove that the anycast nature of query routing (i.e., the number of qualified experts may be more than one) can lead to decentralized search being highly efficient under a wide range of network settings, which is completely different from prior works on the small-world phenomenon.

\subsection{Summary of Contributions}

We study, for the first time, decentralized search in expert networks from the perspective of fundamental performance quantifications. Our contributions are seven-fold:

\emph{1)} We build mathematical models to formulate abundant expert connections to similar experts and a few connections to dissimilar experts, in terms of their expertise differences.\looseness=-1

\emph{2)} To capture the two properties in \emph{1)}, we propose two expert network models: (i) unified model, where all experts have the same overall problem solving abilities, but specialize in different areas, and (ii) diversified model, where experts may have different per-area expertise or different overall problem solving abilities. In the diversified model, the per-expert total problem solving ability exhibits a Gaussian-like distribution as the number of solvable subjects in the network increases.

\emph{3)} We prove that decentralized search is highly efficient under a wide range of network settings for both unified and diversified models; the corresponding average routing path length is between $O(\log n)$ and $O(\log^2 n)$ ($n$: total number of experts).\looseness=-1

\emph{4)} We further establish conditions for the case when decentralized search is ineffective, and develop the corresponding lower bounds to quantify its performance.

\emph{5)} We discuss how above theoretical results are related to the special characteristics of small-world phenomenon in expert networks. We reveal that the existence of small-world phenomenon (under wide conditions) directly leads to efficient decentralized search in expert networks. However, in point-to-point (unicast) networks, one and only one of these conditions enables efficient local-information-based search.

\emph{6)} We demonstrate that decentralized search is robust in the case where experts experience interpretation errors regarding the expertise requirement in the received queries.

\emph{7)} We show that the above theoretical bounds can also approximate the routing performance in real datasets, even though the network structures in these real datasets may not rigorously respect the proposed network models; therefore, these theoretical results can provide guidance in planning practical complex expert networks.

Note that in this paper, the focus is the fundamental understanding of query routing behaviors under the assumption that experts' expertise and social connections are fixed. We acknowledge that both professional and social knowledge of experts may improve over time, thus benefiting the query routing efficiency. In such case where improved routing information is available, the above results remain valid as long as the network parameters in the newly formed expert network (e.g., experts' improved expertise and richer inter-connections) are retrieved and updated (see the case study in Section~\ref{Sect:EvaluationResults}).

\vspace{.2em}
The remainder of the paper is organized as follows. Section~\ref{Sect:ProblemFormulation} formulates the problem. Two models for expert networks are proposed in Section~\ref{Sect:NetworkModel}. Main results of decentralized search in expert networks are presented and analyzed in Section~\ref{Sect:mainResults}, where the corresponding proofs are shown in Section~\ref{Sect:TheoreticalProof}. Experiments are conducted under both synthetic networks and real datasets in Section~\ref{Sect:EvaluationResults}. Finally, Section~\ref{Sect:Conclusion} concludes the paper.

\section{Problem Formulation}
\label{Sect:ProblemFormulation}

In this section, we propose mathematical models to capture expert inter-connections, and then formally present decentralized search and state our research objective.

\subsection{Expert Inter-Connections}
\label{subsect:ExpertConnections}

We assume that in an expert network with $n$ experts, experts can collectively solve problems in up to $m$ different areas.
For all experts, we assume that their expertise in different areas are quantifiable to non-negative integers,
and thereby each expert is associated with an expertise vector defined as follows: The \emph{expertise vector} of expert $u$, denoted by $\textbf{e}^{(u)}$, is an $m\times 1$ column vector with the value in entry $i$ (i.e., $e^{(u)}_i$) indicating $u$'s skill in area $i$ (larger value corresponds to superior skill); $e^{(u)}_j=0$ if $u$ does not have any skill in area $j$. We call $||\textbf{e}^{(u)}||_1:=\sum_i|e^{(u)}_i|$ the \emph{total ability} of expert $u$. Using this concept, we can compare the expertise levels in different areas for one individual expert or in the same area across multiple experts. Furthermore, we define the \emph{expertise distance} from expert $u$ to expert $w$ as $d(u\rightarrow w):=\sum_{i}\max(e^{(w)}_i-e^{(u)}_i,0)$. Intuitively, expertise distance characterizes the superior skills of one expert against another, and it implies that generally $d(u\rightarrow w)\neq d(w\rightarrow u)$. With all these concepts, we are ready to model homophily and heterophily of expert inter-connections.\looseness=-1

\textbf{\emph{Homophily}} refers to the tendency that each expert is rich in connections to peer experts with similar expertise. To characterize such expertise similarity, both inferior and superior skills should be considered when comparing two experts, i.e., expertise difference in all areas between two experts must be within a threshold. Therefore, a natural way to model homophily is: For a universal constant integer $\delta\geq 1$, called \emph{similarity degree}, each expert (denoted by $u$) connects to all experts in set $R:=\{w\in V\setminus \{u\} : ||\textbf{e}^{(w)}-\textbf{e}^{(u)}||_1 \leq \delta\}$, where $V$ is the set of experts in the entire network and ``$\setminus$'' is set minus. All experts in set $R$ are called \emph{local contact}s of expert $u$. When experts are connected by the homophily rule (adding local contacts for each expert), an expert network basis, called \emph{network substrate}, is formed. The network substrate is determined purely by the constant parameter $\delta$, and thus the network substrate does not exhibit any randomness (assuming that the expertise vectors of all experts are fixed). 

\begin{table}[tb]
\renewcommand{\arraystretch}{1.3}
\caption{Notations} \label{t notion}
\footnotesize
\vspace{+.5em}
\centering
\begin{tabular}{r|m{6.3cm}}
  \hline
  \textbf{Symbol} & \textbf{Meaning} \\
  \hline
  $V$, $n$ & set/number of experts ($n=|V|$) \\
  \hline
  $r$   & long-range contact follows an inverse $r$-th power distribution  \\
  \hline
  $C_u$  & set of candidates who can become long-range contacts of $u$    \\
  \hline
  $k$   &  maximum number of long-range contacts for each expert \\
  \hline
  $h$   &  number of rows in the unified model \\
  \hline
  $m$   &  number of elements in an expertise vector  \\
  \hline
  $\textbf{e}^{(u)}$   &  expertise vector ($m\times 1$ column vector) of expert $u$ \\
  \hline
  $(i,\tau)$   &  query in problem area $i$ with difficulty level being $\tau$ (query $(i,\tau)$ is solvable by an expert $u$ with $e_i^{(u)}\geq \tau$) \\
  \hline
  $\lambda$  &  maximum expertise level in each area for any expert in the diversified model     \\
  \hline
  $\delta$  & similarity degree \\
  \hline
  $c$ & scaling factor in the standard deviation of query interpretation: $\sigma=c(\tau-e_i^{(u)})$ for query $(i,\tau)$ at expert $u$\\
  \hline
  $d(u\rightarrow w)$ &  $d(u\rightarrow w)=\sum_{i}\max(e^{(w)}_i-e^{(u)}_i,0)$ expertise distance from $u$ to $w$  \\
  \hline

\end{tabular}
\end{table}
\normalsize

\textbf{\emph{Heterophily}} refers to the phenomenon that each expert has a few connections to experts with fairly dissimilar expertise. These dissimilar experts are called \emph{long-range contact}s, which are crucial in determining the network diameter (average length of the shortest path connecting each pair of nodes). However, unlike homophily, there exists randomness in observing which two dissimilar experts are connected. Therefore, we use a statistical model to capture heterophily, which consists of two steps: (i) for expert $u$, compute the set of candidates $C_u$ that can be long-range contacts of $u$ by $C_u:=\{w\in V\setminus \{u\}: \textbf{e}^{(w)} \npreceq \textbf{e}^{(u)} \}$; (ii) for a universal constant integer $k\geq 1$ and constant $r\geq 0$, expert $u$ has $k$ out-going edges to connect to long-range contacts with independent probabilities.
Specifically, for the $k$ out-going edges from $u$, each edge terminates at expert $w$ ($w\in C_u$) with probability, denoted by $\prob (u\rightarrow w)$, proportional to $[d(u\rightarrow w)]^{-r}$. In other words, $\prob (u\rightarrow w) = [d(u\rightarrow w)]^{-r}/\sum_{v\in C_u}[d(u\rightarrow v)]^{-r}$, called the \emph{inverse $r$-th power distribution}.\looseness=-1

This heterophily model captures that in practical expert networks, expert $u$ connecting to a long-range expert $w$ is because $w$ has superior skills in certain areas compared to $u$, instead of how inferior skills that $w$ exhibits in other areas; therefore, we use expertise distance $d(u\rightarrow w)$ to capture such expertise dissimilarity.
Note that the long-range contact construction rule indicates that the $i$-th and $j$-th edges from $u$ may terminate at the same expert (may even be one of $u$'s local contacts); therefore, $k$ is the maximum number of long-range contacts for each expert.
Moreover, $r$ serves as a structural parameter that controls the scale of long-range contacts for each individual expert. In particular, when $r=0$, all experts in the candidate set $C_u$ are equally likely to be long-range contacts of $u$, which corresponds to a purely random case; when $r$ increases, long-range contacts of an expert tend to only exist within her vicinity (measured by the expertise distance); when $r$ approaches $+\infty$, all long-range contacts disappear, i.e., there is no heterophily in the expert network. In this paper, we study how these parameters relate to the efficiency of decentralized search. All notations used in this paper are summarized in Table~\ref{t notion}.

\begin{figure}[tb]
\centering
\begin{minipage}{.45\linewidth}
  \centerline{\includegraphics[width=1.1\columnwidth]{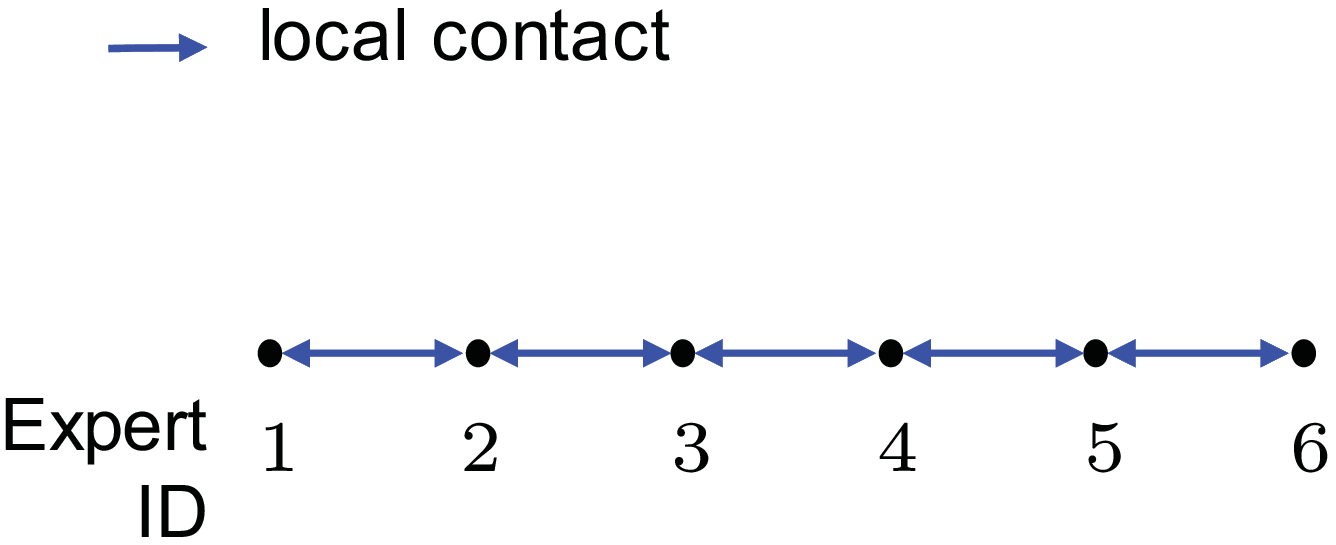}}
  \centerline{\footnotesize (a)}\label{fig:Illustrative1}
\end{minipage}\hfill
\begin{minipage}{.45\linewidth}
  \centerline{\includegraphics[width=1.1\columnwidth]{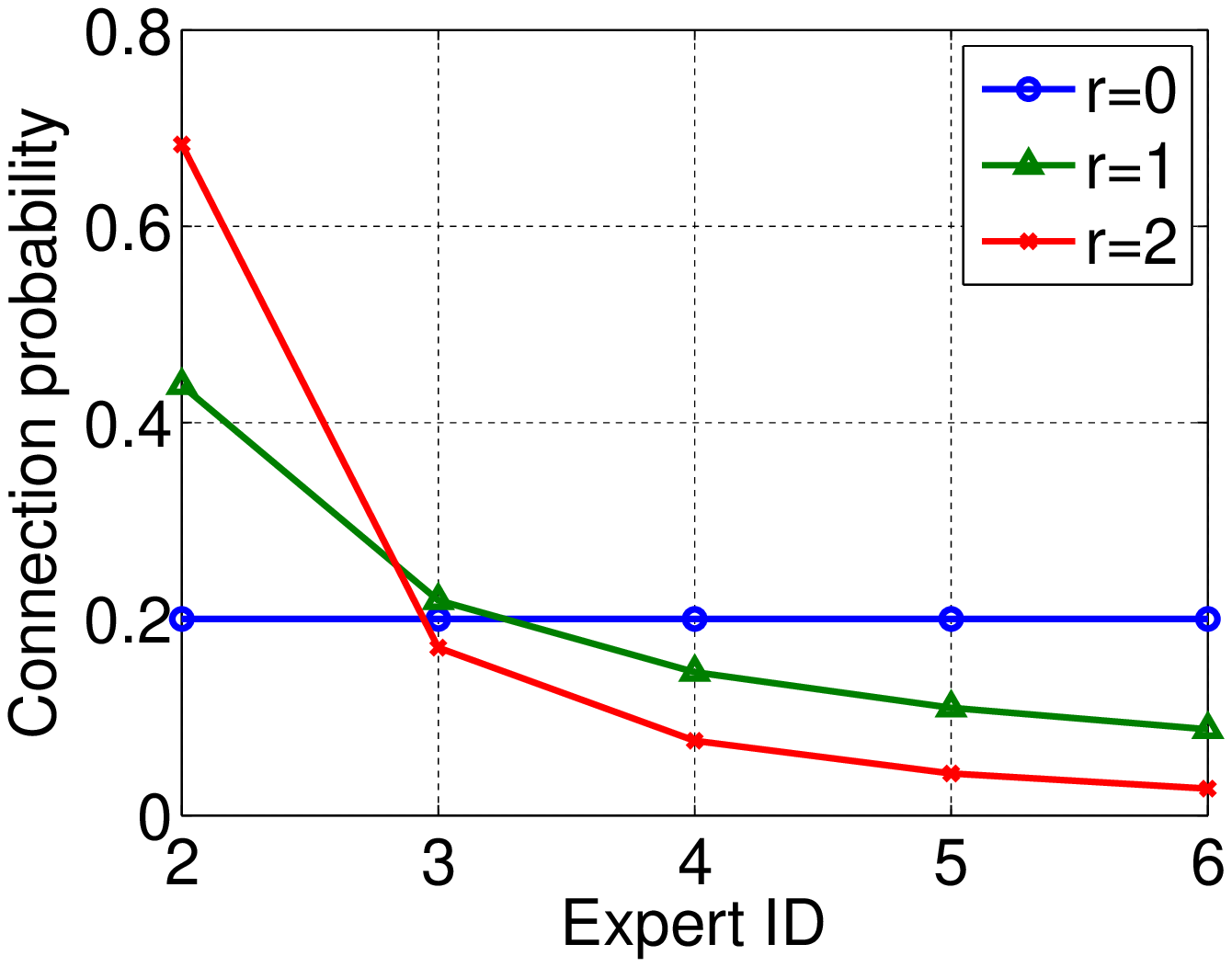}}
  \centerline{\footnotesize (b)}\label{fig:Illustrative2}
\end{minipage}
\caption{Illustrative example: (a) sample network (expertise distance from Expert $i$ to Expert $i+j$ is $j$); (b) probability of Expert $1$ having other experts in (a) as long-range contacts under different values of $r$.\looseness=-1} \label{fig:Illustrative_Example}
\end{figure}

\textbf{Illustrative Example:}
Fig.~\ref{fig:Illustrative_Example}~(a) shows a sample network with six experts, where the expertise distance between two experts is also the difference of their IDs. In Fig.~\ref{fig:Illustrative_Example}~(a), each expert has the most similar experts as local contacts. The probability of Expert $1$ having long-range contacts with other experts is sketched in Fig.~\ref{fig:Illustrative_Example}~(b), which shows that when $r$ increases, the impact of expertise distances becomes salient, thus causing decreased connection probability. 

\subsection{Decentralized Search}

\begin{algorithm}[tb]
\label{Alg:decentralizedRouting}
\small
\SetKwInOut{Input}{input}\SetKwInOut{Output}{output}
\Input{Expert network, query $(i,\tau)$, first query holder $u$}
\Output{Routing path $\mathcal{P}$ for resolving query $(i,\tau)$}
$\mathcal{P}\leftarrow u$\tcp*{"$\leftarrow$":assignment operation}
\While {$e_i^{(u)}-\tau<0$\label{DRM:searchCondition}}
{
$u=\arg\max_{w\in \mathcal{N}(u)}\big(\min(e_i^{(w)}-\tau,0)\big)$ \tcp*{$\mathcal{N}(u)$: set of all (local and long-range) contacts of $u$}\label{DRM:bestNextHop}
$\mathcal{P}\leftarrow\mathcal{P}+u$ \tcp*{append $u$ to $\mathcal{P}$}
}\label{DRM:searchEnd}
\caption{Decentralized Search}
\end{algorithm}
\normalsize

We now formally present decentralized search for query routing. In an expert network, its constituent experts can generally resolve queries in more than one problem area; however, for the queries posted to the network, the most frequent case is that each of them belongs to one and only one problem area\footnote{If a query contains problems in $p$ ($p>1$) areas, then this query can be treated as $p$ separate queries.}. In this regard, we model each query as a $2$-tuple $(i,\tau)$, where $i$ is the problem area to which this query belongs and $\tau$ ($\tau>0$) indicates the corresponding difficulty level, i.e., query $(i,\tau)$ is solvable by experts with expertise level in area $i$ being at least $\tau$. We assume that there is no ambiguity in determining the problem areas of queries, and there exist qualified experts in the network to solve each query, i.e., for any query $(i,\tau)$, $\exists$ expert $w$ with expertise in area $i$ and $e^{(w)}_i\geq \tau$. In this paper, the most crucial assumption is that
for each query holder, besides knowing the expertise vector and (local and long-range) contacts of herself, she also knows the expertise vectors of all her (local and long-range) contacts; however, she does not have knowledge of expertise vectors or contacts of other experts, i.e., no experts have the global view\footnote{If the global picture of the network is fully known to each expert, then simple breadth-first search will suffice to find the shortest query routing path, which is not of interest to this paper.} of the network. Under these assumptions, decentralized search is detailed in Algorithm~\ref{Alg:decentralizedRouting}. In Algorithm~\ref{Alg:decentralizedRouting}, for a given query, if the current query holder $u$ cannot solve this query, then line~\ref{DRM:bestNextHop} searches for the best expert from all $u$'s contacts (with ties broken arbitrarily) as the next routing step. This process continues until a qualified expert is found.


\emph{Remark:} In decentralized search, if a query holder's contacts cannot solve the received query, then this query holder has no knowledge of where the qualified experts are. Therefore, one may concern that the condition in line~\ref{DRM:searchCondition} may never be satisfied for some queries, thus resulting in endless loops. We will show in Section~\ref{Sect:NetworkModel} that the structural properties of the two representative network models abstracted from real networks ensure that at least one expert satisfying the condition in line~\ref{DRM:searchCondition} can be found by Algorithm~\ref{Alg:decentralizedRouting} (although the resulting routing path may be long).

\subsection{Objective}

Suppose that the problem area and the difficulty level in each query are independently and uniformly distributed (subject to the maximum problem solving ability in the network) and the first query holder is also randomly chosen.
Our goal is to understand decentralized search in expert networks by computing its upper/lower bound of the \emph{average routing path length} (measured by the number of hops) under different network structures and expert inter-connections.

\section{Expert Network Models}
\label{Sect:NetworkModel}

Based on expert inter-connection models in Section~\ref{Sect:ProblemFormulation}, we now present two expert network models, i.e., unified and diversified models, which differ by the distribution of expert total abilities. The significance of these models is that each of them captures unique features in real expert networks.\looseness=-1

\subsection{Unified Model}
\label{subsec:unifiedModel}

\begin{figure}[tb]
\centering
\includegraphics[width=3in]{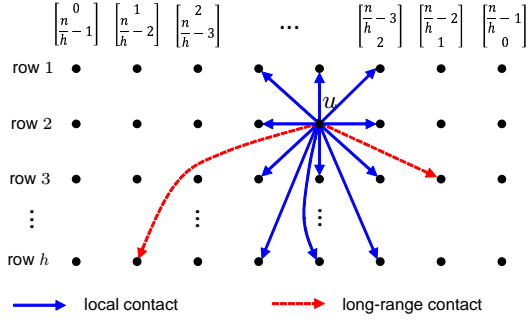}
\caption{Unified model ($\delta=2$, $k=2$, $[\cdot]:$ expertise vector).} \label{Fig:UnifiedModel}
\end{figure}

The first network model is called the \emph{unified model}, where all experts have the same total abilities, i.e., the sum of all entries in the expertise vector of any expert is a constant, though expertise in different areas may vary. The unified model captures the main features in real expert networks where experts have (almost) the same problem solving abilities (e.g., within the same organization or with similar payment), yet also have their own specialties (e.g., they are hired according to company's multi-development goals). Suppose the expert network can solve queries in up to two specific areas (i.e., $m=2$), then the unified model is structured as follows:
\begin{enumerate}
  \item $n$ experts are distributed in an $h\times \frac{n}{h}$ grid ($h$ rows and ${n}/{h}$ columns as shown in Fig.~\ref{Fig:UnifiedModel}, assuming ${n}/{h}$ is an integer);
  \item experts in the same column have the same expertise vector;
  \item in each row, expertise in the first area increases from $0$ to $\frac{n}{h}-1$ by $1$ at each expert in the direction from left to right, while the expertise in the second area increases from $0$ to $\frac{n}{h}-1$ by $1$ at each expert in the opposite direction;\looseness=-1
  \item w.r.t. each expert in a row, she has the most similar experts (i.e., similarity degree $\delta=2$) as local contacts. Thus, each expert with expertise vector $[{0},{(n/h)-1}]^T$ or $[{(n/h)-1},{0}]^T$ has $2h-1$ local contacts; all other experts each has $3h-1$ local contacts (see Fig.~\ref{Fig:UnifiedModel}).\looseness=-1
\end{enumerate}
Then based on the above network substrate, long-range contacts are constructed following the inverse $r$-th power distribution (see Section~\ref{Sect:ProblemFormulation}) for each expert; see Fig.~\ref{Fig:UnifiedModel}.

\emph{Discussions:} This unified model can be extended to cases where $m>2$ (i.e., the network can solve problems in more than two areas). One extension is to model the entire network as a combination of special subnetworks, each specializing in two different areas, and thus each subnetwork in principle can be characterized by the above unified model under $m=2$. Detailed analysis and other possible extensions are omitted due to space limitations. We point out that the significance of the unified model presented in this paper is that it serves as the fundamental building block for complex models with experts all having the same total abilities.\looseness=-1

\subsection{Diversified Model}
\label{subsec:diversifiedModel}

\begin{figure}[tb]
\centering
\includegraphics[width=3.2in]{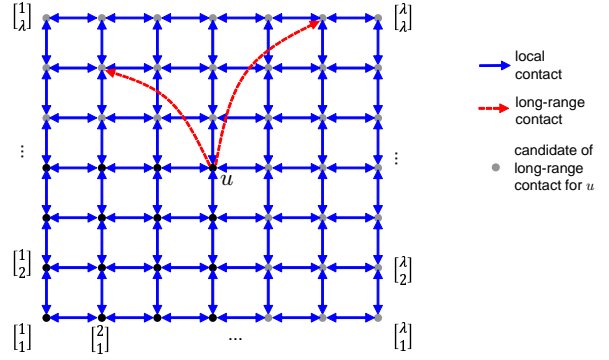}
\caption{Diversified model ($\delta=1$, $m=2$, $k=2$, $[\cdot]:$ expertise vector).} \label{Fig:DiversifiedModel}
\end{figure}

The second network model is called the \emph{diversified model}, in which both the total abilities and specialties may vary for different experts. One advantage of this model is that it naturally captures the Gaussian-like distribution of expert total abilities in real expert networks. Suppose up to $m$ ($m\geq 1$) specific areas can be solved in the expert network. Let $\lambda$ denote the maximum expertise level in each area (i.e., the maximum value for every entry in any expertise vector), then the diversified model is structured as follows:
\begin{enumerate}
  \item suppose for any expertise vector $\textbf{e}$, $\forall i$, $e_i$ is an integer between $1$ and $\lambda$ (i.e., $[1,1,\ldots,1]^T\preceq\textbf{e}\preceq[\lambda,\lambda,\ldots,\lambda]^T$), and each possible value of $\textbf{e}$ corresponds to one expert. Hence, the total number of experts is $n=\lambda^m$;\looseness=-1
  \item each expert only has the most similar experts (i.e., similarity degree $\delta= 1$) as local contacts, thus forming an $m$-dimensional grid (Fig.~\ref{Fig:DiversifiedModel} illustrates a sample $2$-dimensional grid). As Fig.~\ref{Fig:DiversifiedModel} shows, if an expert is not at the grid boundary, then she has $2m$ local contacts; otherwise, the number of local contacts is between $m$ and $2m$.\looseness=-1
\end{enumerate}
Then again based on the above network substrate, long-range contacts are constructed following the inverse $r$-th power distribution for each expert (see the example in Fig.~\ref{Fig:DiversifiedModel}).

\begin{figure}[tb]
\centering
\includegraphics[width=2.5in]{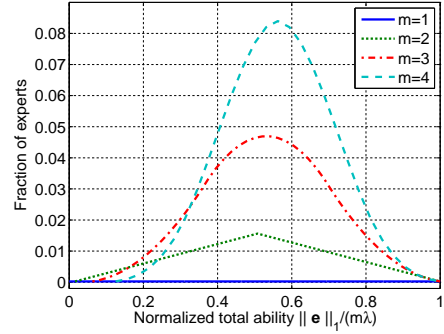}
\caption{Total ability distribution in the diversified model ($n=4096$).} \label{Fig:ExpertiseDistribution}
\end{figure}

\emph{Discussions:} Unlike the unified model, in the diversified model, the total abilities across different experts may be different. In particular, the number of experts with the total ability $\phi$ is
$\sum_{q=0}^{\min(m,(\phi-m)/\lambda)}[(-1)^q\binom{m}{q}\binom{\phi-1-q\lambda}{m-1}]$, and the expected value of total ability is $(m+m\sqrt[m]{n})/2$. By these numerical expressions, the distribution of total abilities is reported in Fig.~\ref{Fig:ExpertiseDistribution} under different values of $m$. The most important property of the diversified model revealed by Fig.~\ref{Fig:ExpertiseDistribution} is that the expert total ability follows a Gaussian-like distribution as $m$ increases. Therefore, the diversified model (when $m>2$) represents a real expert network that is abundant in experts with average problem solving abilities, while lacks experts with significantly superior/inferior total abilities.

\vspace{.5em}
\emph{Remark:} The theoretical results in this paper are based on the network models proposed in this section, where specific expertise distributions over experts are required. Nevertheless, we point out that even if such requirement is not strictly satisfied, our theoretical results still demonstrate high accuracy in predicting query routing performance (see the case study of real datasets in Section~\ref{Sect:EvaluationResults}), thus making contributions from both theoretical and practical perspectives.

\section{Efficiency of Decentralized \\Search in Expert Networks}
\label{Sect:mainResults}

Recall that we assume (in Section~\ref{Sect:ProblemFormulation}) that the problem area and the difficulty level in queries are generated uniformly (up to the maximum problem solving ability per area in the expert network) at random, and the first query holder is also arbitrarily chosen. We now present the corresponding quantitative performance bounds and analysis of decentralized search under the network models proposed in Section~\ref{Sect:NetworkModel}. We then discuss how the small-world phenomenon is related to decentralized search in expert networks. Complete theoretical proofs are presented in Section~\ref{Sect:TheoreticalProof}.

\subsection{Statement of Main Results}
\label{subsec:performanceBounds}
Under the unified and diversified models, we have the performance bounds
of decentralized search stated as below, where $\ln (\cdot)$ denotes natural logarithm. In these results, we assume that $n$ is sufficiently large such that $n>8h$ in the unified model, $\sqrt[m]{n}>8$ in the diversified model, and $n>>k$ in both models (these assumptions are used to avoid the trivial cases where the first query holders are already very close to the destinations even without long-range contacts; see the theorem proofs in Section~\ref{Sect:TheoreticalProof} for details).

\begin{theorem}
\label{thm:monotonicity}
The average routing path length generated by decentralized search is monotonically increasing with $r$ in both unified and diversified models.
\end{theorem}

\begin{theorem}
\label{thm:upperBoundUM}
The average routing path length generated by decentralized search in the unified model is upper bounded by
\begin{align*}
O\Big(\big(\ln\frac{n}{h}\big)^{r+1}\Big) \ \ \mbox{for}\ 0\leq r \leq 1.
\end{align*}
\end{theorem}

\begin{theorem}
\label{thm:lowerBoundUM}
The average routing path length generated by decentralized search in the unified model is lower bounded by
\begin{align*}
\Omega\Big(\frac{1}{k^{{1}/{r}}}\cdot\big(\frac{n}{h}\big)^{\frac{r-1}{r}}\Big) \ \ \mbox{for}\ r> 1.
\end{align*}
\end{theorem}

\begin{theorem}
\label{thm:upperBoundDM}
The average routing path length generated by decentralized search in the diversified model is upper bounded by
\begin{align*}
O\Big(\frac{1}{m^{r}}\cdot\big(\ln n\big)^{r+1}\Big) \ \ &\mbox{for}\ 0\leq r \leq 1.
\end{align*}
\end{theorem}

\begin{theorem}
\label{thm:lowerBoundDM}
The average routing path length generated by decentralized search in the diversified model is lower bounded by
\begin{align*}
\Omega\left(\frac{1}{k^{{1}/{r}}}\cdot n^{\frac{r-1}{mr}}\right) \ \ \mbox{for}\ r> 1.
\end{align*}
\end{theorem}

\begin{corollary}
\label{coro:maxLength}
Any routing path length generated by decentralized search is upper bounded by ${n}/{h}$ in the unified model, and $\sqrt[m]{n}$ in the diversified model.
\end{corollary}

\subsection{Performance of Decentralized Search in Expert Networks}
\subsubsection{$0\leq r\leq 1$}
Theorems~\ref{thm:upperBoundUM} and \ref{thm:upperBoundDM} show that when $0\leq r\leq 1$, the average routing path length using decentralized search is upper bounded by a polylogarithmic function, i.e., a polynomial function of $\ln n$. Therefore, decentralized search is highly efficient in expert searching when $0\leq r\leq 1$. Such high efficiency occurs mainly for the following two reasons: (i) In expert networks, experts exhibit a certain level of randomness in inter-connections, which causes the formation of a network gradient that drives the query to the destination via decentralized search. (ii) Qualified experts for a given query may not be unique, i.e., query routing terminates at any expert who is capable of resolving this query; therefore, the query routing problem in expert networks is an \emph{anycast} problem. Hence, if a query is routed to an over-qualified expert, then this query does not need to be further routed to the expert with the exact required knowledge level as the problem is already solvable.
Another significant insight revealed by Theorems~\ref{thm:upperBoundUM} and \ref{thm:upperBoundDM} is that decentralized search is most effective when $r=0$ (i.e., long-range contacts are randomly selected) as the number of experts $n$ is sufficiently large according to our assumption (i.e., $\ln (n/h)>1$ in the unified model and $(\ln n)/m>1$ in the diversified model); the corresponding routing path length is only logarithmic in the network size.

\subsubsection{$r>1$}
When $r$ increases, Theorem~\ref{thm:monotonicity} shows that the average routing path length also rises. When $r>1$, as shown in Theorems~\ref{thm:lowerBoundUM} and \ref{thm:lowerBoundDM}, the average routing path length can no longer be expressed as a polylogarithmic function, which indicates the ineffectiveness of decentralized search.
Nevertheless, Corollary~\ref{coro:maxLength} proves that for any values of $r$, there is always an upper bound, which is determined by the network size. Therefore, the worst performance of decentralized search happens when $r$ approaches $+\infty$, for which the lower bound is $\Omega(n/h)$ for the unified model, and $\Omega(\sqrt[m]{n})$ for the diversified model. Comparing to Corollary~\ref{coro:maxLength}, this result suggests that the performance bounds in Theorems~\ref{thm:lowerBoundUM} and \ref{thm:lowerBoundDM} are tight when $r$ is large. On the other hand, Theorems~\ref{thm:lowerBoundUM} and \ref{thm:lowerBoundDM} also show that although a larger $k$ may increase the probability of connecting to qualified experts, the impact of $k$ in reducing the average routing path length is weakened when $r>1$.
This is because when $r$ is large, long-range contacts tends to only exist in the vicinity of each expert or even overlap with local contacts, and thus the contribution of a large $k$ diminishes.

\subsection{Uniqueness of Small-World Phenomenon in Expert Networks}
As shown in \cite{Bollobas82,Bollobas88SIAMA,Kleinberg06ICM}, the basic standard for justifying networks with small-world phenomenon is that the associated network diameters can be expressed as a polylogarithmic function of $n$. Thus, by Theorems~\ref{thm:upperBoundUM} and \ref{thm:upperBoundDM}, we conclude that not only does the small-world phenomenon \emph{exist} in expert networks under \emph{both} unified and diversified models when $0\leq r\leq 1$, but also decentralized search is able to find these short paths. This is in sharp contrast with the unicast problem in prior works on the small-world phenomenon. Specifically, assuming that individual connections also follow the inverse $r$-th power distribution (constructed based on their lattice distances) in the unicast problem, \cite{Kleinberg00STOC} shows that though the small-world phenomenon is pervasive for a range of $r$, decentralized search is \emph{only} efficient under a unique value of $r$. This is because
in a unicast problem with the destination being $t$, along the way from the current message holder to $t$, if one intermediate node routes this message to a long-distant node (corresponding to one over-qualified expert in expert networks) that is beyond $t$, then this message needs to be routed back to $t$, thus resulting in longer routing paths.

\section{Theoretical Proofs}
\label{Sect:TheoreticalProof}

To prove the theorems in Section~\ref{Sect:mainResults}, let $(i,\tau)$ denote a query posted to the network, and $s$ the first holder of $(i,\tau)$. We define $L$, a random variable, as the total number of hops spent on finding a qualified expert for $(i,\tau)$ (i.e., starting from $s$ and terminating at any expert that can solve $(i,\tau)$). Then it suffices to determine the expected value $\mathbb{E}[L]$ under different network settings.


\subsection{Proof of Theorem~\ref{thm:monotonicity}}


Regarding query $(i,\tau)$, there exists at least one qualified expert according to our assumption. Let $T$ be the set of all experts qualified to solve $(i,\tau)$. We sort all experts in $T$ in an increasing order w.r.t. their expertise in area $i$ such that $e_i^{(t_1)}\leq e_i^{(t_2)}\leq\ldots\leq e_i^{(t_\zeta)}$, where $T=\{t_1,t_2,\ldots,t_\zeta\}$. For all other experts, let $U$ denote the set containing all experts with expertise better than $s$ but less than $t_1$ in area $i$. Then when delivering $(i,\tau)$ from $s$ to $T$, only experts in $U$ can become the relay experts. Furthermore, from $s$ to $T$, at each relay expert $w$, $e^{(t_1)}_i-e^{(w)}_i$ is strictly decreasing, as each expert in both unified and diversified models has at least one (local or long-range) contact who has better expertise in area $i$. Therefore, we can divide this routing process (before reaching any expert in $T$) into several phases, where in phase $j$, $e^{(t_1)}_i-e^{(w)}_i$ w.r.t. the current query holder $w$ is greater than $j\Delta$ and at most $(j+1)\Delta$ ($\Delta$ is a constant). Then suppose the query routing process is currently in phase $j$, then the probability of leaving phase $j$ in the next routing step is proportional to the probability of experts in phase $j$ having long-range contacts to experts in phase $q$ ($q<j$) or experts in $T$ (note that the normalization factor in the inverse $r$-th power distribution w.r.t. an expert is fixed as it only relies on the underlying network structure). Moreover, the probability of having such long-range contacts is strictly increasing with $1/r$ according to the inverse $r$-th power distribution. Thus, if experts associated with phase $j$ are employed as relay experts, then the average number of hops in phase $j$ is monotonically increasing with $r$. We note that by the definition of phase, the number of hops in certain phases can be zero. Nevertheless, the probability of experiencing a certain phase, say phase $j$, is also strictly increasing with $r$. Thus, the overall average number of hops in phase $j$ remains monotonically increasing with $r$. As $\mathbb{E}[L|(i,\tau)]$ for the given query $(i,\tau)$ is the sum of average number of hops in each phase, we have that $\mathbb{E}[L|(i,\tau)]$ is monotonically increasing with $r$. The above argument remains valid regarding all other queries, we therefore conclude that the total expectation $\mathbb{E}[L]$ is monotonically increasing with $r$ for both the unified and diversified model.
\hfill$\blacksquare$

\subsection{Proof of Theorem~\ref{thm:upperBoundUM}}
To prove Theorem~\ref{thm:upperBoundUM}, we first prove that the following inequality holds.
\small
\begin{equation}\label{eq:Jensen}
  \frac{d^{-r}}{\sum^{n'}_{j=1}j^{-r}}\geq \frac{1}{n'(\sum^{n'}_{j=1}j^{-1})^r}\mbox{ for } 0\leq r\leq 1,\mbox{ and }1\leq d\leq n'.
\end{equation}
\normalsize
Let $g(x):=x^r$, where $0\leq r\leq 1$. Define random variable $y$ as follows: $y=1/j$ with probability $1/n'$ for $j=1,2,\ldots,n'$. Then we have
\small
\begin{align*}
g(\mathbb{E}[y])=\Big(\frac{\sum^{n'}_{j=1}j^{-1}}{n'}\Big)^r,\
\mathbb{E}[g(y)]=\frac{\sum^{n'}_{j=1}j^{-r}}{n'}.
\end{align*}
\normalsize
Since $g(x)$ is a concave function, by Jensen's theorem \cite{Boyd_Convex}, the inequality $g[\mathbb{E}(x)]\geq \mathbb{E}[g(x)]$ holds. Therefore,
\small
\begin{align*}
n'\big(\sum^{n'}_{j=1}j^{-1}\big)^r\geq (n')^r \sum^{n'}_{j=1}j^{-r}\geq d^r \sum^{n'}_{j=1}j^{-r},
\end{align*}
\normalsize
as $1\leq d\leq n'$. Thus, (\ref{eq:Jensen}) is correct.

Now we can prove Theorem~\ref{thm:upperBoundUM}. In the unified model, starting from one expert $u$, there are at most $2h$ experts with the expertise distance being $j$ ($j$ is an integer). Thus, we have
\small
\begin{equation}\label{eq:normalizationUM}
\sum_{w\in C_u}[d(u\rightarrow w)]^{-r}\leq \sum_{j=1}^{(n/h)-1}2h j^{-r}.
\end{equation}
\normalsize
According to the distribution of long-range contacts, $\prob (u\rightarrow w) = [d(u\rightarrow w)]^{-r}/\sum_{v\in C_u}[d(u\rightarrow v)]^{-r}$; therefore, by (\ref{eq:normalizationUM}),
\small
\begin{equation}\label{eq:UMprobLower}
\prob (u\rightarrow w) \geq \frac{[d(u\rightarrow w)]^{-r}}{\sum_{j=1}^{(n/h)-1}2h j^{-r}}
\end{equation}
\normalsize
Since $d(u\rightarrow w)\leq (n/h)-1$ and $0\leq r\leq 1$, by (\ref{eq:Jensen}), (\ref{eq:UMprobLower}) is further lower bounded by
\small
\begin{align*}
\frac{[d(u\rightarrow w)]^{-r}}{\sum_{j=1}^{(n/h)-1}2h j^{-r}}&\geq \frac{1}{2h(\frac{n}{h}-1)\big(\sum_{j=1}^{(n/h)-1}j^{-1}\big)^r}\\
&\geq \frac{1}{2(n-h)\big(1+\ln \frac{n}{h}\big)^r}\geq \frac{1}{2(n-h)\big(\ln \frac{3n}{h}\big)^r}.
\end{align*}
\normalsize
Thus, $\prob (u\rightarrow w)\geq [2(n-h)(\ln \frac{3n}{h})^r]^{-1}$. For the received query $(i,\tau)$, suppose there are $\eta$ qualified experts in each row of the unified model. Then together there are $\eta h$ experts, denoted by set $T$, capable of solving $(i,\tau)$. Let $\prob (u\rightarrow T)$ denote the probability that $u$ has a contact in $T$ when there are $k$ out-going edges, then
\small
$
\prob (u\rightarrow T)\geq\sum_{w\in T} \prob (u\rightarrow w)\geq {\eta h }[{2(n-h)(\ln \frac{3n}{h})^r}]^{-1}
$\normalsize
. Let $\Gamma:={\eta h }[{2(n-h)(\ln \frac{3n}{h})^r}]^{-1}$ and $Y_{\eta}$ denote the total number of hops that are spent for solving $(i,\tau)$ when there are $\eta h$ qualified experts. We have
\small
\begin{equation}\label{eq:EYetaI}
\begin{aligned}
\mathbb{E}[Y_{\eta}] &= \sum^{\infty}_{j=1}\prob[Y_{\eta}\geq j]\leq \sum^{\infty}_{j=1}\Big(1-\Gamma\Big)^{j-1}=\frac{1}{\Gamma}.
\end{aligned}
\end{equation}
\normalsize
Recall that $L$ denotes the total number of hops spent for solving $(i,\tau)$. Since $i$ and $\tau$ in $(i,\tau)$ are uniformly distributed ($i=\{1,\ldots,m\}$, $0< \tau\leq (n/h)-1$), we can derive
\small
\begin{equation}\label{eq:EY}
\begin{aligned}
\mathbb{E}[L]&=\sum^{(n/h)-1}_{\eta=1} \frac{1}{(n/h)-1} \mathbb{E}[Y_{\eta}]\\
&\leq \frac{2(n-h)\big(\ln \frac{3n}{h}\big)^r}{ h }\cdot\frac{h}{n-h}\cdot\sum^{(n/h)-1}_{\eta=1} \eta^{-1}\\
&\leq {2\big(\ln \frac{3n}{h}\big)^r}(1+\ln \frac{n}{h})\leq {2\big(\ln \frac{3n}{h}\big)^{r+1}}.
\end{aligned}
\end{equation}
\normalsize
Therefore, the average routing path length $\mathbb{E}[L]$ is upper bounded by $O\big((\ln\frac{n}{h})^{r+1}\big)$, when $0\leq r\leq 1$.
\hfill$\blacksquare$

\subsection{Proof of Theorem~\ref{thm:lowerBoundUM}}
\label{subsec:proofThm3}
In the unified model, for any expert $u$, $\sum_{w\in C_u}[d(u\rightarrow w)]^{-r}$ is lower bounded by a constant $\xi$ ($\xi$ is the cardinality of set $S:=\{w\in V\setminus \{u\}:d(u\rightarrow w)=1,e^{(w)}_i-e^{(u)}_i=1\}$).
Then the probability that $u$ has a long-range contact $w$ with $e^{(w)}_i-e^{(u)}_i$ greater than $l$, denoted by $\prob[e^{(w)}_i-e^{(u)}_i>l]$, is\looseness=-1
\small
\begin{equation}\label{eq:probDistance>l}
\begin{aligned}
\prob[e^{(w)}_i-e^{(u)}_i>l] &\leq \frac{\sum_{j=l+1}^{(n/h)-1}\xi j^{-r}}{\xi}\leq \int_{l}^\infty x^{-r} dx=\frac{l^{1-r}}{r-1}.
\end{aligned}
\end{equation}
\normalsize
Then following similar arguments in \cite{Kleinberg00STOC}, we define $\alpha:={1}/{r}$, $\beta:={(r-1)}/{r}$, and $\theta:={\min(r-1,1)}/{(8k)}$. Let $T$ denote the set of all experts who can solve $(i,\tau)$ ($|T|\geq 1$). We define $\mathcal{A}_j$ to be the event that in the $j$-th hop since decentralized search starts query routing, query $(i,\tau)$ reaches an expert $w$ ($w\notin T$) that has a long-range contact $v$ with $e^{(v)}_i-e^{(w)}_i\geq (kn/h)^\alpha$. Then let $\mathcal{A}:=\bigcup_{1\leq j\leq \theta (kn/h)^\beta} \mathcal{A}_j$ denote the event that this occurs in the first $\theta (kn/h)^\beta$ hops. As the probability of a union of events is upper bounded by the sum of their individual probabilities, we have
\small
\begin{equation}\label{eq:UMprobA}
\begin{aligned}
\prob [\mathcal{A}] &\leq \sum_{1\leq j\leq \theta (kn/h)^\beta} \prob[\mathcal{A}_j]\leq \theta (kn/h)^\beta \frac{k(kn/h)^{\alpha-\alpha r}}{r-1}
\leq\frac{1}{8}.
\end{aligned}
\end{equation}
\normalsize
Let $t_1$ be the expert in $T$ with the minimum expertise in solving $(i,\tau)$. Define $\mathcal{B}$ to be the event that $e^{(t_1)}_i-e^{(s)}_i> \frac{n}{8h}$ (recall that $s$ denotes the first query holder). Since $n/h>8$ according to our assumption, we have
\small
\begin{align*}
\prob[\mathcal{B}]=\sum^{\frac{n}{h}-\frac{n}{8h}}_{x=1}\frac{1}{(n/h)-1}\cdot\frac{\frac{n}{h}-x-\frac{n}{8h}}{n/h}> {1}/{3}.
\end{align*}
\normalsize
By (\ref{eq:UMprobA}), $\prob[\overline{\mathcal{B}} \vee \mathcal{A}]< \frac{2}{3}+\frac{1}{8}$; therefore, $\prob [\mathcal{B}\wedge\overline{\mathcal{A}}]> \frac{5}{24}$.

We now prove that if $\mathcal{B}$ occurs and $\mathcal{A}$ does not occur, then $\mathbb{E}[L|\mathcal{B}\wedge \overline{\mathcal{A}}]\geq \theta (kn/h)^\beta$ as follows: If $\mathcal{A}$ does not occur, then at each routing step, query $(i,\tau)$ can move toward $T$ by a distance of at most $(kn/h)^\alpha$ in area $i$; then after $\theta (kn/h)^\beta$ steps, the total moved distance in area $i$ is at most
\small
\begin{align*}
\theta (kn/h)^{\alpha+\beta}=\theta k (n/h)={\min(r-1,1)n}/{(8h)}\leq\frac{n}{8h},
\end{align*}
\normalsize
i.e., $(i,\tau)$ cannot reach any expert in $T$ if $\mathcal{B}$ occurs and $\mathcal{A}$ does not occur. Therefore, $\mathbb{E}[L|\mathcal{B}\wedge \overline{\mathcal{A}}]\geq \theta (kn/h)^\beta$. Accordingly,
\small
\begin{align*}
\mathbb{E}[L]&\geq \mathbb{E}[L|\mathcal{B}\wedge \overline{\mathcal{A}}]\cdot \prob[\mathcal{B}\wedge \overline{\mathcal{A}}]\\
&> \frac{5}{24}\theta (kn/h)^\beta=\frac{5\min(r-1,1)\cdot k^{-\frac{1}{r}}}{192}\big(\frac{n}{h}\big)^{\frac{r-1}{r}}.
\end{align*}
\normalsize
Therefore, when $r>1$, the average routing path length in the unified model is lower bounded by
$
\Omega( k^{-\frac{1}{r}}(\frac{n}{h})^{\frac{r-1}{r}}) $.
\hfill$\blacksquare$

\subsection{Proof of Theorem~\ref{thm:upperBoundDM}}

We follow similar arguments as those in the proof of Theorem~\ref{thm:upperBoundUM}. In the diversified model, there are up to $c_0\lambda^{m-1}$ ($c_0$: a constant) experts with the same expertise distance from any expert; therefore, under the diversified model, (\ref{eq:normalizationUM}) in the unified model is changed to
$
\sum_{w\in C_u}[d(u\rightarrow w)]^{-r}\leq \sum_{j=1}^{m\lambda-m}c_0\lambda^{m-1} j^{-r}
$
in the diversified model. Since $d(u\rightarrow w)\leq m\lambda-m$, (\ref{eq:UMprobLower}) becomes
\small
\begin{align*}
\prob (u\rightarrow w) &\geq \frac{[d(u\rightarrow w)]^{-r}}{\sum_{j=1}^{m\lambda-m}c_0\lambda^{m-1} j^{-r}}\\
\mbox{by (\ref{eq:Jensen}), }&\geq \frac{1}{mc_0\lambda^{m}\big(\sum_{j=1}^{m\lambda-m}j^{-1}\big)^r}\geq \frac{1}{mc_0\lambda^{m}\big(\ln {(3m\lambda)}\big)^r}.
\end{align*}
\normalsize

To compute $\mathbb{E}[L]$ for query $(i,\tau)$, let $\eta:=\lambda-\tau+1$. Then there are $\eta \lambda^{m-1}$ qualified experts, denoted by set $T$. Let $\prob (u\rightarrow T)$ be the probability that $u$ has a long-range contact in $T$, then
\small
$
\prob (u\rightarrow T)\geq\eta \lambda^{m-1}\sum_{w\in T} \prob (u\rightarrow w)\geq {\eta}{(c_0 m\lambda)^{-1}}\big(\ln {(3m\lambda)}\big)^{-r}.
$\normalsize
Then following similar arguments for computing (\ref{eq:EYetaI}--\ref{eq:EY}) in the unified model, we have
$
\mathbb{E}[L]\leq{c_0m\left(\ln(3m\lambda)\right)^{r+1}}
$.
Thus, when $0\leq r\leq 1$, $\mathbb{E}[L]$ is upper bounded by
\small$O({m^{-r}(\ln n)^{r+1}})$\normalsize,
where we use $\lambda^m=n$.\looseness=-1
\hfill$\blacksquare$

\subsection{Proof of Theorem~\ref{thm:lowerBoundDM}}
Let $T$ denote the set of all experts who can solve query $(i,\tau)$. Similar to the proof of Theorem~\ref{thm:lowerBoundUM}, in the diversified model, $\sum_{w\in C_u}[d(u\rightarrow w)]^{-r}$ is lower bounded by a constant $\xi$ (defined in Section~\ref{subsec:proofThm3}); therefore, similar to (\ref{eq:probDistance>l}), we have
\small
$
\prob[e^{(w)}_i-e^{(u)}_i>l] \leq ({\sum_{j=l+1}^{m\lambda-m}\xi j^{-r}})/{\xi}\leq{l^{1-r}}/{(r-1)}.
$\normalsize\
Again, define $\alpha:=\frac{1}{r}$, $\beta:=\frac{r-1}{r}$, and $\theta:=\frac{\min(r-1,1)}{8k}$. Define $\mathcal{A}_j$ to be the event that in the $j$-th hop, query $(i,\tau)$ reaches an expert $w$ ($w\notin T$) that has a long-range contact $v$ with $e^{(v)}_i-e^{(w)}_i\geq (k\lambda)^\alpha$. Then let $\mathcal{A}:=\bigcup_{1\leq j\leq \theta (k\lambda)^\beta} \mathcal{A}_j$ denote the event that this occurs in the first $\theta (k\lambda)^\beta$ hops. Thus,
$
\prob [\mathcal{A}] \leq \sum_{1\leq j\leq \theta (k\lambda)^\beta} \prob[\mathcal{A}_j]\leq{1}/{8}.
$

Next, similar to the proof of Theorem~\ref{thm:lowerBoundUM}, let $t_1$ be the expert in $T$ with the minimum expertise in solving $(i,\tau)$, and $\mathcal{B}$ the event that  $e^{(t_1)}_i-e^{(s)}_i> \frac{\lambda}{8}$ (recall $s$ is the first query holder). Since $\lambda=\sqrt[m]{n}>8$ according to our assumption, we again have $\prob[\mathcal{B}]> \frac{1}{3}$ and $\prob [\mathcal{B}\wedge\overline{\mathcal{A}}]> \frac{5}{24}$.
Then using similar arguments for proving Theorem~\ref{thm:lowerBoundUM}, we have $\mathbb{E}[L|\mathcal{B}\wedge \overline{\mathcal{A}}]\geq \theta (k\lambda)^\beta$. Hence,
$
\mathbb{E}[L]\geq \mathbb{E}[L|\mathcal{B}\wedge \overline{\mathcal{A}}]\cdot \prob[\mathcal{B}\wedge \overline{\mathcal{A}}]>{\rho k^{-\frac{1}{r}}}\lambda^{\frac{r-1}{r}}
$, where $\rho:={5\min(r-1,1)}/{192}$.
Since $\lambda=\sqrt[m]{n}$, the lower bound in the diversified model is $\Omega(k^{-\frac{1}{r}} n^{\frac{r-1}{mr}})$ when $r>1$.\looseness=-1
\hfill$\blacksquare$

\subsection{Proof of Corollary~\ref{coro:maxLength}}
The statement in Corollary~\ref{coro:maxLength} follows by considering the worst routing case: Query $(i,\tau)$ is of the maximum difficulty level (i.e., $\tau =\max_{u\in V} e_i^{(u)}$), and the expert with the worst expertise in area $i$ is selected as the first query holder.
\hfill$\blacksquare$

\section{Experiments}
\label{Sect:EvaluationResults}


\begin{figure}[tb]
\centering
\begin{minipage}{0.75\linewidth}
  \centerline{\includegraphics[width=1.1\columnwidth]{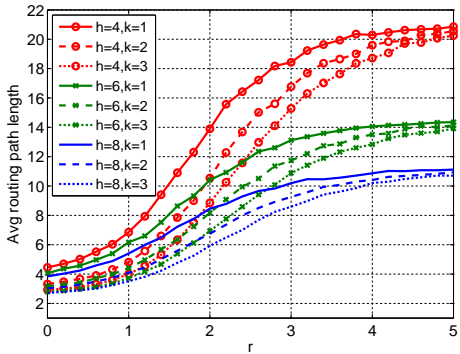}}
  \centerline{\footnotesize (a) Unified model ($n=240$)}
\end{minipage}\hfill
\begin{minipage}{0.75\linewidth}
  \centerline{\includegraphics[width=1.1\columnwidth]{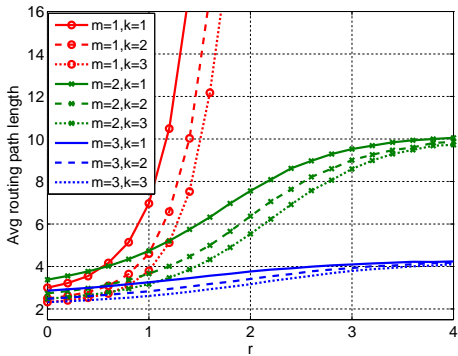}}
  \centerline{\footnotesize (b) Diversified model ($n=729$)}
\end{minipage}
\caption{Average routing path length ($100$ random network realizations for each network parameter setting, $500$ Monte Carlo runs per network realization).}\label{Fig:Model_results}
\end{figure}

In this section, we evaluate the performance of decentralized search by computing its average routing path length under the unified/diversified model and explaining the corresponding observations using the performance bounds in Section~\ref{Sect:mainResults}. Then, we study the robustness of decentralized search against query interpretation errors. Finally, we compare the predicted query routing time using the performance bounds to the actual routing time in a case study of real datasets for justifying the applicability of the performance bounds to real networks.

\begin{figure}[tb]
\centering
\begin{minipage}{0.75\linewidth}
  \centerline{\includegraphics[width=1.1\columnwidth]{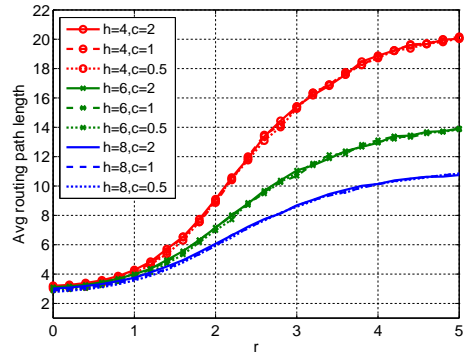}}
  \centerline{\footnotesize (a) Unified model ($n=240$)}
\end{minipage}\hfill
\begin{minipage}{0.75\linewidth}
  \centerline{\includegraphics[width=1.1\columnwidth]{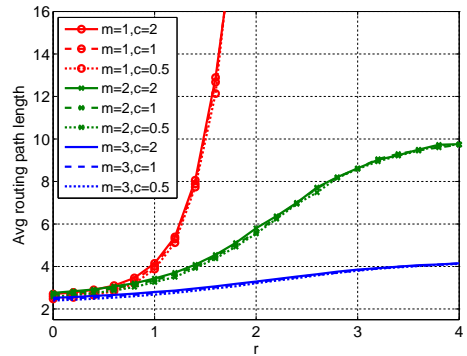}}
  \centerline{\footnotesize (b) Diversified model ($n=729$)}
\end{minipage}
\caption{Average routing path length with query interpretation errors ($k=3$, $100$ random network realizations for each network parameter setting, $500$ Monte Carlo runs per network realization).}\label{Fig:results_with_errors}
\end{figure}

\emph{\textbf{Performance Under the Unified/Diversified Model:}} To evaluate the performance of decentralized search, we select $n=240$ and $h=4,6,8$ for the unified model, and $n=729$ and $m=1,2,3$ for the diversified model, both under $k=1,2,3$. We generate queries by randomly selecting the corresponding problem area and the required expertise level (subject to the network-wise maximum problem solving capability) in both the unified and diversified model; moreover, the first query holders are also randomly chosen. For each network parameter setting, $100$ random network realizations are generated, and $500$ Monte Carlo runs (each run corresponds to a newly generated query that is randomly distributed to an expert as the first query holder) are conducted on each network realization. Using decentralized search, the resulting routing path length averaged over all network realizations and Monte Carlo runs are reported in Fig.~\ref{Fig:Model_results}~(a) (unified model) and Fig.~\ref{Fig:Model_results}~(b) (diversified model).

In Fig.~\ref{Fig:Model_results}, as expected, we first observe that the average routing path length increases with $r$ (supported by Theorem~\ref{thm:monotonicity}). The most significant conclusion we can draw from Fig.~\ref{Fig:Model_results} is that they confirm the high efficiency of decentralized search when $0\leq r\leq 1$ for both network models (as proved in Theorems~\ref{thm:upperBoundUM} and \ref{thm:upperBoundDM}). Specifically, compared to the network size ($240$ and $729$ experts in the unified and diversified models), decentralized search achieves extremely small routing path length, i.e., between $2$ to $5$ when $r=0$ and $3$ to $8$ when $r=1$. Moreover, Fig.~\ref{Fig:Model_results} demonstrates that the performance of decentralized search is similar under different network parameters/models when $r$ is small (e.g., around $0$). However, when $r$ increases, the performance under different parameters begins to depart, and the performance under small $h$ (or $m$) degrades significantly when $r>1$, for which Theorems~\ref{thm:lowerBoundUM} and \ref{thm:lowerBoundDM} provide quantitative bounds to capture such performance deterioration.
Nevertheless, such performance degradation converges when $r$ is large, because all routing path lengths are constrained by the upper bound (independent of $r$) established in Corollary~\ref{coro:maxLength}. Moreover, Fig.~\ref{Fig:Model_results} shows that the benefit of a larger number of long-range contacts is not obvious, especially in the case when $r$ is large. This is because, as shown in Theorems~\ref{thm:lowerBoundUM} and \ref{thm:lowerBoundDM}, for each expert, having a single contact with significantly different specialties (controlled by $r$) is more effective than having multiple contacts (controlled by $k$) with only limited expertise dissimilarity in reducing the average routing path length.\looseness=-1

\emph{\textbf{Robustness of Decentralized Search:}} Thus far, we assume that each expert has accurate estimation of problem difficulties; in other words, the expertise levels required by the queries in different areas are perceived exactly the same as the ground truth for all experts in the network. However, this may not always be true. Therefore, we study how the performance of decentralized search is affected by misinterpreting query difficulties as follows: When attempting to solve query $(i,\tau)$, if expert $u$ has sufficient expertise in area $i$ (i.e., $e_i^{(u)}\geq\tau$), then she solves $(i,\tau)$; otherwise, her estimation of the expertise required for solving $(i,\tau)$ follows the truncated Gaussian distribution with minimum value $\tau_{\min}$, mean $\mu$, and standard deviation $\sigma$ being functions of $u$ and $(i,\tau)$. In particular, $\tau_{\min}=:e_i^{(u)}$, $\mu:=\tau$, and $\sigma:=c(\tau-e_i^{(u)})$, where $c$ is a scaling factor. She then uses this estimated value
$\tau'$ to find the best next hop from her contacts following the decentralized search rule. In this estimation error model, we capture the fact that in real networks, query estimation accuracy increases as the expert expertise gets closer to the actual requirement. The results of decentralized search under different levels (controlled by the scaling factor $c$) of such estimation errors averaged over multiple Monte Carlo runs are shown in Fig.~\ref{Fig:results_with_errors}~(a) (unified model) and Fig.~\ref{Fig:results_with_errors}~(b) (diversified model). These results confirm that under query estimation errors, the performance of decentralized search remains stable; therefore, decentralized search is a robust and reliable solution for query routing in expert networks.\looseness=-1

\begin{figure}[tb]
\centering
\begin{minipage}{0.5\linewidth}
  \centerline{\includegraphics[width=1.1\columnwidth]{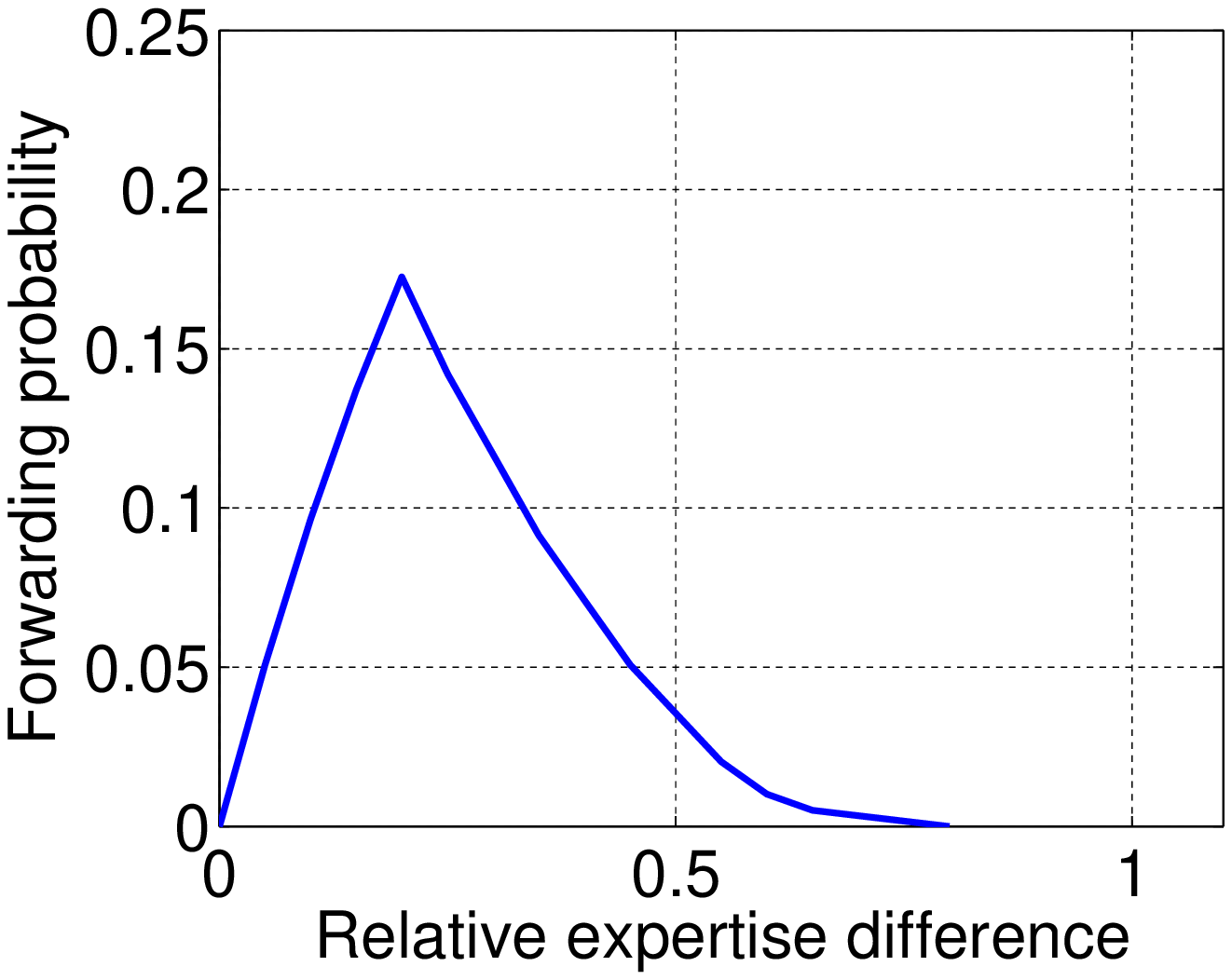}}
  \centerline{\footnotesize (a)}
\end{minipage}\hfill
\begin{minipage}{0.5\linewidth}
  \centerline{\includegraphics[width=1.1\columnwidth]{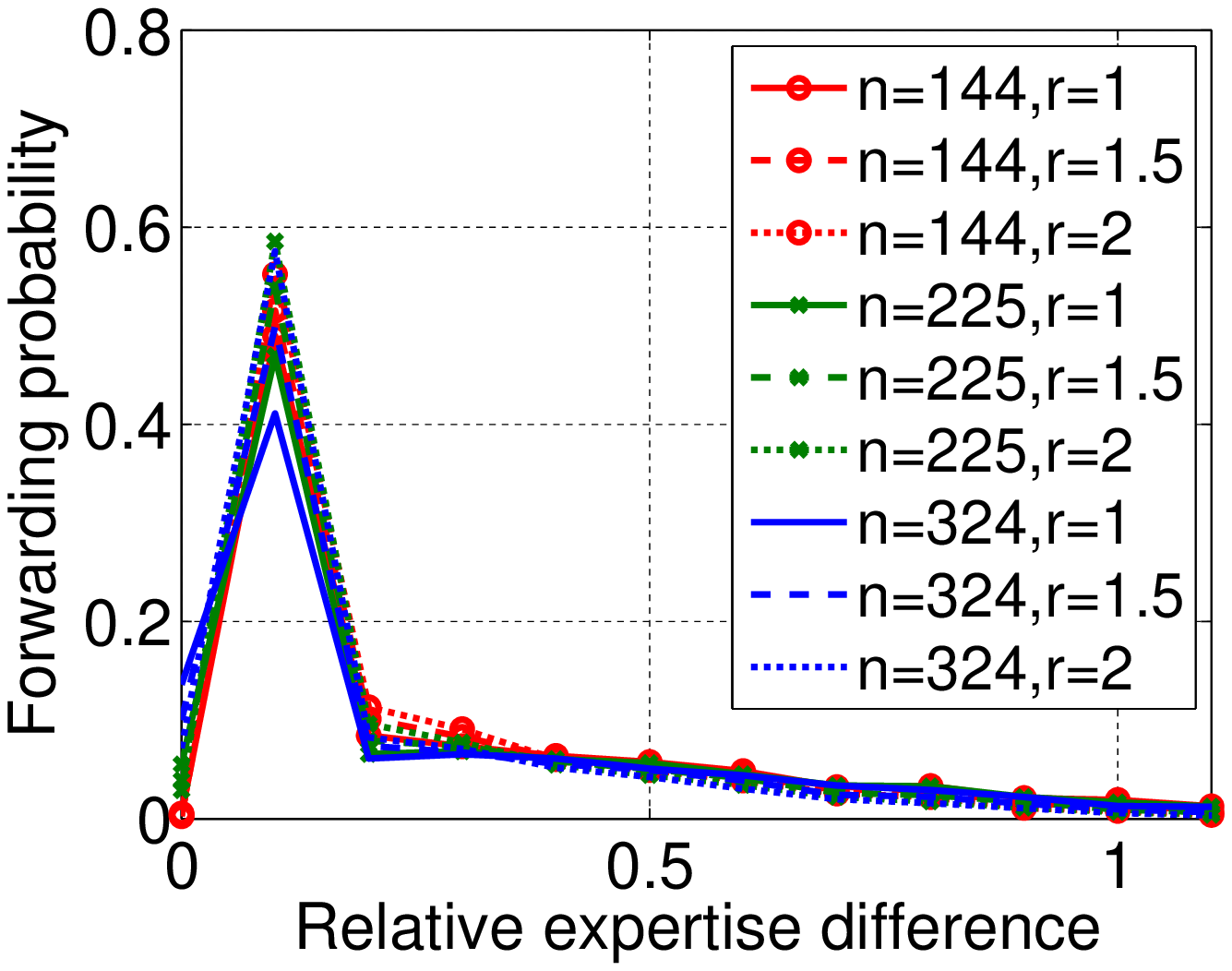}}
  \centerline{\footnotesize (b)}
\end{minipage}
\caption{(a) Query routing behavior in real datasets; (b) query routing behavior of decentralized search in the diversified model ($m=2$, $k=1$, $100$ random network realizations for each network parameter setting, $500$ Monte Carlo runs per network realization).}\label{Fig:Forwarding_Probability}
\end{figure}

\emph{\textbf{Case Study of Real Datasets:}} Next, using our theoretical results, we analyze real-world query routing data collected from the IT service department of one of Fortune 500 companies throughout 2006. Depending on query contents, these datasets are categorized into four independent classes: Operating System 1 (OS-1), Operating System 2 (OS-2), Database, and Web Service. Table~\ref{t RealData} lists the network parameters in these datasets. For these datasets, we first observe that their network structures can be characterized by the diversified model\footnote{Note that in Table~\ref{t RealData}, the expert network associated with each dataset is not strictly structured as that in Fig.~\ref{Fig:DiversifiedModel}, i.e., some experts in Fig.~\ref{Fig:DiversifiedModel} may be missing; however, the performance prediction remains accurate.} for the case of $m=2$ and $k=1$. In Table~\ref{t RealData}, two values of $r$ (i.e., $r_1$ and $r_2$) are derived from these datasets based on expert inter-connections, where $r_2$ corresponds to new connections after applying a \emph{mentoring program} to the original expert networks (associated with $r_1$). In this mentoring program, some less-skilled experts are mentored by experienced experts, which equivalently reduces the value of $r$. In order to apply our theoretical results in Section~\ref{Sect:mainResults}, we still need to justify if the query routing behaviors in these real datasets share any similarities with decentralized search. To this end, we define relative expertise difference as $||\textbf{e}^{(w)}-\textbf{e}^{(u)}||_1/||\textbf{e}^{(u)}||_1$, where $w$ is the next hop expert selected by expert $u$. The query forwarding probability versus relative expertise difference averaged over all queries received in the four networks of the datasets is shown in Fig.~\ref{Fig:Forwarding_Probability}~(a). As comparison, we also compute the same metric based on decentralized search in the diversified model as reported in Fig.~\ref{Fig:Forwarding_Probability}~(b), where three networks that of similar network sizes and similar values of $r_1$ and $r_2$ as those in the real datasets are evaluated. We note that Fig.~\ref{Fig:Forwarding_Probability}~(a) and (b) have similar shapes, i.e., the expert with neither too similar nor too different expertise is selected with high probability as the next hop; therefore, the routing behaviors in these datasets do exhibit a certain level of decentralized search. Hence, we can use our results in Section~\ref{Sect:mainResults} on decentralized search to predict the routing performance in these real datasets. Let $\overline{L}_i$ denote the average query routing path length under $r_i$ ($i=1,2$). We compare the real $\overline{L}_1/\overline{L}_2$ with the predicted $\overline{L}_1/\overline{L}_2$ using Theorem~\ref{thm:lowerBoundDM} (as $r_1,r_2>1$ for all datasets). The comparison in Table~\ref{t RealData} shows that using the theoretical performance bounds, the predicted routing path length is accurate (the error is $21.1\%$ for Database, and $5.9\%\sim 13.2\%$ for other datasets); therefore, the theoretical results in this paper can naturally serve as an efficient tool for analyzing/predicting behaviors in real expert networks. Moreover, these datasets also suggest that to achieve high routing efficiency, network owners can take proactive actions to adjust the expert connections such that the resulting network condition ($r_2$ is close to $1$) approaches the high efficiency region (i.e., $0\leq r\leq 1$).

\begin{table}[tb]
\renewcommand{\arraystretch}{1.3}
\caption{ Performance in Real Expert Networks \normalfont(captured by the diversified model using decentralized search, where $m=2$ and $k=1$. $\overline{L}_i$: average query routing path length under $r_i$, $i=1,2$)} \label{t RealData}
\vspace{+.2em}
\centering
\begin{tabular}{c|c|c|c|c|c}
\hline
\multirow{2}{*}{Datasets} &  \multirow{2}{*}{${n}$}&   \multirow{2}{*}{$r_1$}   &   \multirow{2}{*}{$r_2$}   & real & predicted  \\
        &       &           &           &   $\overline{L}_1/\overline{L}_2$ & $\overline{L}_1/\overline{L}_2$\\
\hline
OS-1    &   184 &   1.98    &   1.44    &   1.44    &   1.63\\
\hline
OS-2    &   122 &   1.24    &   1.04    &   1.31    &   1.45\\
\hline
Database     &   305 &   1.61    &   1.01    &   2.37    &   2.87\\
\hline
Web Service&  266 &   1.14    &   1.04    &   1.19    &   1.26\\
\hline		
\end{tabular}
\end{table}

\section{Conclusion}
\label{Sect:Conclusion}
We investigated the efficiency of local-information-based decentralized search for query answering in expert networks, focusing on quantifying the performance of decentralized search under various network settings. Incorporating common expert social inter-connection tendencies, we proposed two expert network models, each representing a unique distribution of expert problem solving abilities in the network. Under these two network models, we established fundamental theories demonstrating when decentralized search is exceptionally effective in finding short query routing paths. In cases where decentralized search is ineffective, we also quantified how the performance deterioration is correlated to network structures. Evaluations and comparisons of these theoretical results in both synthetic networks and real datasets confirm the efficiency/robustness of decentralized search in expert networks as well as the significance of the developed performance bounds in guiding real network design.

\bibliographystyle{IEEEtran}
\bibliography{mybibSimplifiedA}

\end{document}